\documentclass[conference]{IEEEtran}
\usepackage{cite}
\usepackage{amsmath,amssymb,amsfonts}
\usepackage{algorithmic}
\usepackage{graphicx}
\usepackage{textcomp}
\usepackage{xcolor}
\usepackage{algorithm}
\usepackage{algorithmic}
\usepackage{bm}
\usepackage{subfigure}
\usepackage{multirow}

\ifCLASSINFOpdf
\else
\fi
\begin{document}
%
\title{Finding Sets of Pareto Sets in Real-World Scenarios \--- A Multitask Multiobjective Perspective}

\author{
\IEEEauthorblockN{Jiao Liu, Yew Soon Ong, Melvin Wong} \\
\IEEEauthorblockA{ 
College of Computing \& Data Science, Nanyang Technological University, Singapore \\
\{jiao.liu, asysong, wong1357\}@ntu.edu.sg}
}

\maketitle

\begin{abstract}
Recently, evolutionary multitasking has been employed to generate a ``\textit{set of Pareto sets}" (SOS) for machine learning models, addressing diverse task settings across heterogeneous environments. This involves creating a repository of compact, specialized solution models that are collectively tailored to each specific task setting and environment, enabling users to select the most suitable model based on particular specifications and preferences. In this paper, we further demonstrate the versatility and applicability of the SOS concept across diverse domains, focusing on three real-world problems: engineering design problems, inventory management problems, and hyperparameter optimization problems. Additionally, as evolutionary multitasking has proven effective in generating the SOS, we investigate the performance of current evolutionary multitasking methods on these real-world problems. Subsequently, we present visualizations of the generated SOS in both decision and objective spaces, complemented by the development of a measurement to gauge the similarity between different Pareto sets corresponding to diverse tasks. Finally, we show that by systematically examining the shifts in Pareto optimal designs across different task settings though the SOS solutions, users can gain deeper understandings on the dynamic interplay between design solutions and their performance in different settings or contexts.


\end{abstract}

\begin{IEEEkeywords}
Evolutionary multitasking, multiobjective optimization, set of Pareto sets
\end{IEEEkeywords}

%
\IEEEpeerreviewmaketitle


\section{Introduction}\label{sec1}


Recently, \textit{evolutionary multitasking} (EMT) has emerged as a prominent research area in the evolutionary computation community~\cite{gupta2017insights}. Unlike traditional evolutionary algorithms, EMT harnesses latent synergies between distinct yet correlated optimization tasks, resulting in superior search performances characterized by enhanced solution quality and convergence~\cite{gupta2015multifactorial,feng2023evolutionary}. Leveraging these advantages, EMT has demonstrated its capability to provide a \textit{set of Pareto sets} (SOS) for machine learning models in a single pass, aiming to address diverse task settings and various resource-constrained environments~\cite{10188456}. The SOS concept involves creating a repository of compact, specialized models tailored to multiple narrowly defined task settings across various environments, naturally conceptualized as a \textit{multitask multiobjective optimization problem}~\cite{9493747,10180214,9925083,ju2023hybrid}. This collective of specialized models offers dynamic scalability, seamlessly adapting to a priori unknown objectives, intentions, and constraints set by human end-users.

While the original SOS is designed for machine learning tasks, we believe it holds immense value when extended to other domains such as engineering~\cite{zhang2014multiobjective,niloy2023brief,avigad2009interactive} and management sciences~\cite{li2022dynamic}. For instance, in the automotive industry, engineers seek to identify various structures with different masses and crashworthiness under diverse load cases~\cite{zhang2014multiobjective}. The SOS concept becomes invaluable by providing a collection of Pareto sets for different load conditions, empowering engineers to conveniently select preferred Pareto solutions based on specific load scenarios. Similarly, in supply chain management, where the need to dynamically adjust inventory based on external environmental changes is critical~\cite{li2022dynamic}, the SOS concept offers a repertoire of Pareto sets for different environments, facilitating informed decision-making in inventory management.


In this paper, our primary objective is to explore the potential of EMT in generating SOSs for real-world problems across diverse domains. Despite the extensive study of EMT approaches in recent years, much of the research has focused on benchmark problems. There has been limited investigation into the performance of current EMT methods in generating SOSs for real-world problems. Our aim is to address this gap by assessing the capability of existing EMT approaches in generating SOSs for three distinct types of real-world problems: engineering design problems~\cite{tanabe2020easy}, inventory management problems~\cite{tsou2008multi}, and hyperparameter optimization problems~\cite{min2020generalizing}. Moreover, we conducted a visualization analysis of the SOS for these problems, illustrating the characteristics of their solution sets and explaining why EMT is well-suited for generating SOS for the employed real-world problems. Additionally, we highlight an advantage of the SOS: it empowers engineers to dissect the impact of environmental features on Pareto sets, thereby fostering a deeper understanding of the characteristics of a specific category of real-world optimization problems.

The structure of this paper is as follows. Section II introduces the basic concepts of the SOS and the related work of EMT. Section III presents the real-world problems used in the study. In Section IV, we conduct experimental studies to explore the capabilities of five different EMT algorithms in generating the SOS. Finally, Section V offers conclusions for this paper.

\section{Background}
\subsection{Formulation of the Set of Pareto Sets}\label{sec2.1}

The SOS represents a collective of Pareto optimal solutions designed to address multiple task settings (e.g., machine learning models for different tasks) while simultaneously adapting to various optimization objectives (e.g., the predicted accuracy and the consumed computational resources of a machine learning model). Moreover, it is crucial to ensure that, for each type of task setting, a corresponding solution with preferred performance on the objectives can be identified from the SOS. Let $f_{k,i}(\cdot)$, where $k \in \{1,\ldots,K\}$ and $i \in \{1,\ldots,m\}$, be the $i$th objective of the solution on the $k$th task setting. (Throughout this paper, it is assumed that, for all these measurements, smaller values indicate better results.) Formulating the identification of the SOS involves framing the problem as the following multitask multiobjective  optimization problem: 
\begin{equation}\label{eqn:sos}
\begin{aligned}
\ & \forall T_k, k \in \{ 1,\ldots,K \}, \\
\min: & \ \ \textbf{F}_k({\textbf{x}}_k) = \left( f_{k,1}({\textbf{x}}_k), \ldots , f_{k,m}({\textbf{x}}_k)\right), \\
\text{s.t.} & \ \ \bm{\textbf{x}}_k \in \Omega_k \subset \mathbb{R}^{d},
\end{aligned}
\end{equation}
where $T_k$ denotes the $k$th task setting, $\textbf{F}_k(\cdot)$ is the objective function vector corresponding to the $k$th task setting, ${\textbf{x}}_k = ({x}_{k,1},\ldots,{x}_{k,d})$ represents the decision vector corresponding to the $k$th task setting, and $\Omega_k$ is the decision space corresponding to the $k$th task setting. Given the formulation in \eqref{eqn:sos}, the associated key concepts~\cite{branke2008multiobjective} are explained as follows:
\begin{itemize}
    \item \textit{Pareto Dominance}: Solution $\textbf{x}_k^{(a)}$ is said to Pareto dominate another solution $\textbf{x}_k^{(b)}$ on the $k$th task setting, if $\forall i \in \{1,2,\ldots,m\}$, $f_{k,i}(\textbf{x}_k^{(a)}) \leq f_{k,i}(\textbf{x}_k^{(b)})$ and $\exists i' \in \{1,2,\ldots,m\}$ such that $f_{k,i'}(\textbf{x}_k^{(a)}) < f_{k,i'}(\textbf{x}_k^{(b)})$.
    \item \textit{Pareto Optimal Solutions}: Solution $\textbf{x}_k^{*}$ is said to be Pareto optimal on the $k$th task setting if there are no other candidate solutions that can dominate $\textbf{x}_k^{*}$.
    \item \textit{Pareto Set}: The Pareto set (PS) consists of all the Pareto optimal solutions.
    \item \textit{Pareto Front}: {The image of the Pareto set in the objective space is referred to as the Pareto front (PF)}
\end{itemize}

The result of \eqref{eqn:sos}, also representing the desired SOS, can be expressed as: 
\begin{equation}\label{eqn:sos_result}
\begin{aligned}
& PS_{k} = \{ \textbf{x}_{k}^{(1)*}, \textbf{x}_{k}^{(2)*}, \textbf{x}_{k}^{(3)*}, \ldots \}, \\
& \mathcal{S} = \bigcup_{k=1}^{K} PS_{k},
\end{aligned}
\end{equation}
where $\textbf{x}^{(\cdot)*}_k$ is a Pareto optimal solution on the $k$th task setting, $PS_k$ is a solution set of the $k$th task setting, and $\mathcal{S}$ is the optimal SOS. 

\subsection{Evolutionary Multitasking}
Evolutionary algorithms are classical methods for solving multiobjective optimization problems~\cite{branke2008multiobjective}. Traditional evolutionary algorithms primarily concentrate on generating sets of non-dominated solutions for a single optimization problem. In contrast, drawing inspiration from multitask learning techniques in the machine learning community, EMT has been proposed to enhance optimization efficiency by leveraging the knowledge embedded in a set of related optimization tasks~\cite{gupta2017insights}. Unlike multitask learning~\cite{zhang2018overview}, which primarily aims to improve predictive accuracy in machine learning models, EMT emphasizes enhancing the optimization process's convergence.

In EMT, one prevalent technique for knowledge transfer involves performing crossover operations on solutions associated with different tasks. Such methods are commonly referred to as implicit transfer~\cite{feng2023evolutionary}. The MFEA algorithm is the most well-known example based on this approach~\cite{gupta2015multifactorial}. Inspired by MFEA, advanced EMT algorithms have been introduced, incorporating strategies like resource allocation~\cite{wen2017parting,gong2019evolutionary} or adaptive knowledge transfer~\cite{bali2019multifactorial,liang2020evolutionary}, to enhance the effective utilization of shared knowledge across diverse tasks. Another set of techniques involves learning the search mapping among tasks, referred to as explicit transfer~\cite{feng2023evolutionary}. Over the past few years, various models, such as autoencoders~\cite{feng2018evolutionary} and kernel-based nonlinear mapping~\cite{10075542}, have been proposed to capture the relationship between different optimization tasks better, thus improving the performance of the evolutionary multitask optimization algorithms.

While numerous studies on EMT have been proposed in recent years, they often focus on evaluating the convergence performance of algorithms on multitask multiobjective benchmarks. In contrast to previous works, this paper primarily explores the capability of EMT in generating SOSs for real-world problems. 

\section{Set of Pareto Sets in Real-World Problems}

In this section, we introduce the real-world problems considered in this paper, including the engineering design problems, the inventory management problems, and the hyperparameter optimization problems.

\subsection{Engineering Design Problems}

\subsubsection{Four Bar Truss Design~\cite{cheng1999generalized}}
\begin{table}[]
\centering
\caption{Parameter Settings of the Four Bar Truss Design Problems.\label{Tab:RE1}}
{\begin{tabular}{c|c|c|c|c|ccc}
\hline
Problem & Tasks & $F$ & $\sigma$ & $L$ & $E$  \\
\hline
    & Task 1 & 10 &  10 & 200 & 2.00E+05 \\
EO1 & Task 2 & 8 &  10 & 200 & 1.50E+05 \\
    & Task 3 & 8 &  8 & 200 & 1.50E+05 \\
\hline
\end{tabular}}
\end{table}

In this problem (recorded as EO1 here), we aim to generate the SOS for the four-bar truss design problem. This design problem contains two minimized criteria: the structural volume and the joint displacement. The former aims to decrease the weight of the entire structure, while the latter aims to enhance the strength of the structure as a whole. Let $f_1$ and $f_2$ be the structural volume and the joint displacement, respectively, the four-bar truss design problem can be defined as follows:
\begin{equation}\label{eqn:re241}
\begin{aligned}
& f_1(\textbf{x}) = L(2x_1 + \sqrt{2}x_2 + \sqrt{x_3} + x_4), \\
& f_2(\textbf{x}) = \frac{FL}{E}(\frac{2}{x_1} + \frac{2\sqrt{2}}{x_2} - \frac{2\sqrt{2}}{x_3} + \frac{2}{x_4}), \\
\end{aligned}
\end{equation}
where $F$ and $L$ determine the load condition and the general structure of the four-bar truss, $E$ represents Young's modulus determined by the materials, and the four variables $x_1,\ldots,x_4$ denote the lengths of the four bars, respectively. The value range for the four decision variables is defined as follows: $x_1, x_4 \in [a,3a]$ and $x_2, x_3 \in [\sqrt{2}a,3a]$, where $a = F/\sigma$ and $\sigma$ denotes the loading pressure.

In traditional multiobjective optimization, the problem is typically solved under specific task settings, where $F$, $L$, $E$, and $\sigma$ are assigned specific values. In this scenario, only the Pareto optimal solutions corresponding to that particular setting can be obtained. This inconveniences engineers, as any changes in load cases or materials necessitate resolving a new multiobjective optimization problem. However, if we can provide a SOS for engineers, where each solution set contains Pareto optimal solutions corresponding to a specific load case and material, engineers would only need to select the desired solution from the relevant solution set. This streamlined process would be more convenient for engineers.

In this paper, we consider generating the SOS for three different task settings. Each task setting has a set of corresponding parameters. The details are listed in Table~\ref{Tab:RE1}.  

\subsubsection{Hatch Cover Design~\cite{amir1989nonlinear}}

Hatch cover design is also a classical engineering design problem. This problem contains two minimized objectives:
\begin{equation}\label{eqn:re224}
\begin{aligned}
& f_1(\textbf{x}) = x_1 + 120 x_2, \\
& f_2(\textbf{x}) = \sum_{i=1}^{4} \max \{ - g_i(\textbf{x}), 0 \}, \\
\end{aligned}
\end{equation}
where 
\begin{equation}\label{eqn:re224_g}
\begin{aligned}
& g_1(\textbf{x}) = 1 - \frac{\sigma_b}{\sigma_{b,max}}, \\
& g_2(\textbf{x}) = 1 - \frac{\tau}{\tau_{max}}, \\
& g_3(\textbf{x}) = 1 - \frac{\delta}{\delta_{max}}, \\
& g_4(\textbf{x}) = 1 - \frac{\sigma_b}{\sigma_k}.
\end{aligned}
\end{equation}
For this design problem, the two decision variables $x_1 \in [ 0.5,4 ]$ and $x_2 \in [4,50]$ represent the flange thickness and the beam height of the hatch cover, respectively. Similar to the first engineering design problem, we create three tasks by setting $E$, $\sigma_{b,max}$ and $\delta_{max}$ to three different settings as shown in Table~\ref{Tab:RE2}, while the other parameters are set as follows: $\tau_{max} = 450 kg/cm$, $\sigma_k = Ex_1^2/100 kg/cm^2$, $\sigma_b = 4500/(x_1 x_2) kg/cm^2$, $\tau = 1800/x_2 kg/cm^2$, and $\delta = 56.2 \times 10^4/(Ex_1x_2^2)$.

\begin{table}[t]
\centering
\caption{Parameter Settings of the Hatch Cover Design Problems.\label{Tab:RE2}}
{\begin{tabular}{c|c|c|c|cccc}
\hline
Problem & Tasks & $E$ & $\sigma_{b,max}$ & $\delta_{max}$ \\
\hline
    & Task 1 & 700000$kg/cm^2$ &  700$kg/cm^2$ & 1.5$cm$ \\
EO2 &Task 2 & 500000$kg/cm^2$ &  700$kg/cm^2$ & 2$cm$ \\
    &Task 3 & 500000$kg/cm^2$ &  500$kg/cm^2$  & 2$cm$ \\
\hline
\end{tabular}}
\end{table}

\subsubsection{Welded Beam Design~\cite{ray2002swarm}}

\begin{table}[t]
\centering
\caption{Parameter Settings of the Welded Beam Design Problems.\label{Tab:RE3}}
{\begin{tabular}{c|c|c|c|cccc}
\hline
Problem & Tasks & $P$ & $L$ & $E$ \\
\hline
    & Task 1 & 6000$lb$ &  14$in$ & 3.00E+07$psi$ \\
EO3 & Task 2 & 4000$lb$ &  14$in$ & 2.00E+07$psi$ \\
    & Task 3 & 4000$lb$ &  10$in$  & 2.00E+07$psi$ \\
\hline
\end{tabular}}
\end{table}

The welded beam design is defined to minimize the following two objectives:
\begin{equation}\label{eqn:re342}
\begin{aligned}
& f_1(\textbf{x}) = 1.10471 x_1^2 x_2 + 0.04811 x_3 x_4 (14 + x_2) + \lambda g(\textbf{x}), \\
& f_2(\textbf{x}) = \frac{4 P L^3}{E x_4 x_3^3} + \lambda g(\textbf{x}), \\
\end{aligned}
\end{equation}
where 
\begin{equation}\label{eqn:re342_g}
\begin{aligned}
& g(\textbf{x}) = \sum_{i=1}^{4} \max \{ - g_i(\textbf{x}), 0 \} \\
& g_1(\textbf{x}) = \tau_{max} - \tau(\textbf{x}), \\
& g_2(\textbf{x}) = \sigma_{max} - \sigma(\textbf{x}), \\
& g_3(\textbf{x}) = x_4 - x_1, \\
& g_4(\textbf{x}) = P_C(\textbf{x}) - P, \\
& \tau(\textbf{x}) = \sqrt{ (\tau')^2 + \frac{2 \tau' \tau'' x_2}{2R} + (\tau'')^2 }, \\
& \tau' = \frac{P}{\sqrt{2} x_1 x_2}, \\
& \tau'' = \frac{MR}{J}, \\
& M = P(L + \frac{x_2}{2}), \\
& R = \sqrt{\frac{x_2^2}{4} + \left( \frac{x_1 + x_3}{2} \right)^2}, \\
& J = 2\left( \sqrt{2} x_1 x_2 \left( \frac{x_2^2}{12} + \left( \frac{x_1 + x_3}{2} \right)^2 \right) \right), \\
& \sigma(\textbf{x}) = \frac{6PL}{x_4 x_3^2}, \\
& P_C(\textbf{x}) = \frac{4.013E \sqrt{x_3^2 x_4^6/36}}{L^2}\left( 1 - \frac{x_3}{2L}\sqrt{\frac{E}{4G}} \right).
\end{aligned}
\end{equation}
In \eqref{eqn:re342} and \eqref{eqn:re342_g}, the four decision variables represent the size of the beam, where $x_1, x_4 \in [0.125,5]$ and $x_2, x_3 \in [0.1,10]$. We also give create three tasks by setting $P$, $L$ and $E$ to different values, as shown in Table~\ref{Tab:RE3}. The remained parameters are set as follows: $G = 12 \times 10^6 pse$, $\tau_{max} = 13600 psi$, $\sigma_{max} = 30000 psi$, and $\lambda = 1000$.

\subsection{Inventory Management Problems}

\begin{table}[t]
\centering
\caption{Parameter Settings of the Inventory Management Problems.\label{Tab:IM}}
\resizebox{8.5cm}{!}{\begin{tabular}{c|c|c|c|c|c|cc}
\hline
Problem & Tasks & $D$ & $\sigma_{L}$ & $r$ & $K$ & $c$ \\
\hline
      & Task 1 & 3412 & 53.354 & 0.26 & 80  & 27.5 \\
IM1   & Task 2 & 490  & 5.027  & 0.3  & 80  & 241  \\
      & Task 3 & 4736 & 57.911 & 0.3  & 135 & 29.41 \\
\hline
      & Task 1 & 4736 & 57.911 & 0.3  & 135 & 29.41 \\
IM2   & Task 2 & 200  & 2.969  & 0.26 & 80 & 233  \\
      & Task 3 & 215  & 2.781  & 0.3  & 80 & 435  \\
\hline
      & Task 1 & 215   & 2.781  & 0.3  & 80 & 435  \\
IM3   & Task 2 & 22774 & 245.333 & 0.26 & 135 & 12.6 \\
      & Task 3 & 10557 & 85.395 & 0.26 & 135 & 2.14 \\
\hline
\end{tabular}}
\end{table}

\renewcommand{\arraystretch}{1.5}
\begin{table*}[t]
\centering
\caption{CHV results of MFEA, MFEA-II, EMT-ET, and NSGA-II averaged over 20 independent runs of every optimizer.}\label{tab:chv}
\begin{tabular}{c|cccccccccccccc}
\hline
\multirow{2}{*}{Problems}  & \multicolumn{1}{c}{MO-MFEA} & \multicolumn{1}{c}{MO-MFEA-II} & EMT-ET  & \multicolumn{1}{c}{NSGA-II} \\
\cline{2-5}
    & \multicolumn{1}{c}{\textit{CHV}$\pm$\textit{Std Dev}} & \multicolumn{1}{c}{\textit{CHV}$\pm$\textit{Std Dev}}  & \multicolumn{1}{c}{\textit{CHV}$\pm$\textit{Std Dev}} & \multicolumn{1}{c}{\textit{CHV}$\pm$\textit{Std Dev}} \\
\hline
EO1 & 2.1659$\pm$0.0015          & \textbf{2.1679$\pm$0.0017}          & 2.1668$\pm$0.0015  & 2.1435$\pm$0.0079 \\
EO2 & 2.3611$\pm$0.0010          & \textbf{2.3625}$\pm$\textbf{0.0009} & 2.3617$\pm$0.0012 & 2.3604$\pm$0.0016 \\
EO3 & \textbf{2.9521}$\pm$\textbf{0.0089} & 2.9321$\pm$0.0238          & {2.9398$\pm$0.0214} & 2.8870$\pm$0.0398 \\
IM1 & 2.8820$\pm$0.0006          & \textbf{2.8826$\pm$0.0007}          & 2.8821$\pm$0.0012 & 2.8818$\pm$0.0016 \\
IM2 & 2.8770$\pm$0.0009          & \textbf{2.8778$\pm$0.0010}          & 2.8771$\pm$0.0011 & 2.8770$\pm$0.0012 \\
IM3 & 2.8791$\pm$0.0007          & \textbf{2.8797$\pm$0.0007}          & 2.8795$\pm$0.0007 & 2.8797$\pm$0.0005 \\
HPO & 1.7510$\pm$0.0046          & \textbf{1.7661}$\pm$\textbf{0.0065} & 1.7458$\pm$0.0086 & 1.7473$\pm$0.0164 \\
\hline
\end{tabular}
\end{table*}

Inventory management is a classical problem in operational research, addressing decisions about when to order and how much to order under different control mechanisms. This paper focuses on the continuous review $(Q,u)$ system, where an order of size $Q$ is placed whenever the inventory position drops to the reorder point $u$. The determination of $(Q, u)$ depends on lead time and demand fluctuations to minimize inventory costs and maximize customer service. Inspired by the model proposed by Agrell \textit{et al}.~\cite{tsou2008multi}, this paper considers the following two objectives:
\begin{equation}\label{eqn:re241}
\begin{aligned}
 f_1(Q,u) = & C(Q,u) = \frac{UD}{Q} + \left( \frac{Q}{2} + u \sigma_L \right)rc, \\
 f_2(Q,u) = & n_s(Q,u) + S(Q,u)  \\
  = & \frac{D}{Q} \int_u^{\infty} \phi(x) \text{d}x + \frac{D\sigma_L}{Q} (\phi(u) - u(1 - \Phi(u))), \\
\end{aligned}
\end{equation}
where $Q \in [\sqrt{2UD/(rc)},D]$, $u \in [1,D/\sigma_L]$, $C(Q,u)$ is the expected total annual cost, $n_s(Q,u)$ is the expected number of stockout occasions annually, $S(Q,u)$ is the expected annual number of items stocked out, $D$ is the expected annual demand, $\sigma_L$ is the standard deviation of lead time demand, $D_L$ is the lead time demand, $u$ is the fixed setup cost, $c$ is the per item cost of manufacture, $r$ is the annual cost of capital, and $\phi(\cdot)$ and $\Phi(\cdot)$ are the probability density function and the cumulative distribution function of a standard Gaussian distribution, respectively.

It is important to note that with changes in the external environment, such as economic conditions or seasonal variations, parameters like the expected annual demand and lead time demand will also change. Consequently, the inventory management problem and its corresponding optimal solutions will vary. If we can offer a SOS containing solution sets for different environments, the decision-maker only needs to select a suitable and preferred solution. This streamlined approach adds significant convenience to the decision-making process. 

In this paper, we try to generate the SOS for three sets of problems, each comprising three optimization tasks. The parameter settings for these tasks are detailed in Table~\ref{Tab:IM}.

\subsection{Hyperparameter Optimization Problems}

Hyperparameter optimization is a fundamental topic in machine learning, where the configuration of hyperparameters not only impacts model performance but also influences the consumption of computational resources. In this paper, we aim to generate a set of hyperparameter sets for machine learning models across a set of related tasks. Taking inspiration from~\cite{10188456}, we partition the MNIST dataset into even and odd numbers, treating them as two subtasks of the handwritten number image classification task. We utilize the classical LeNet-5 architecture~\cite{lecun1998gradient} for these tasks and optimize four hyperparameters: the learning rate of the optimizer, and the number of channels of the first, second, and third convolutional layer, respectively. The learning rate is within the range of $[0,0.1]$, and the number of channels of all the convolutional layers is within the range of $\{1,\ldots,32\}$. The classification accuracy, considered as a maximization objective, and the number of parameters, considered as a minimization objective, serve as the two objectives. Similar to the concept introduced in~\cite{10188456}, such a SOS allows the decision-maker to select suitable hyperparameters tailored to specific task settings and resource-constrained environments.

\begin{figure*}[!t]
	\begin{center}
		\subfigure[]{\label{re241_dec}\includegraphics[width=0.65\columnwidth]{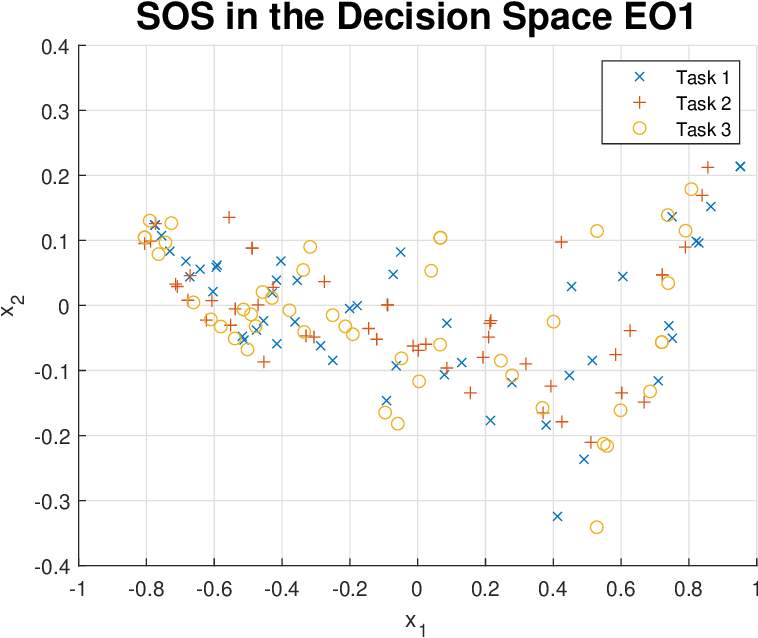}}
		\subfigure[]{\label{re224_dec}\includegraphics[width=0.65\columnwidth]{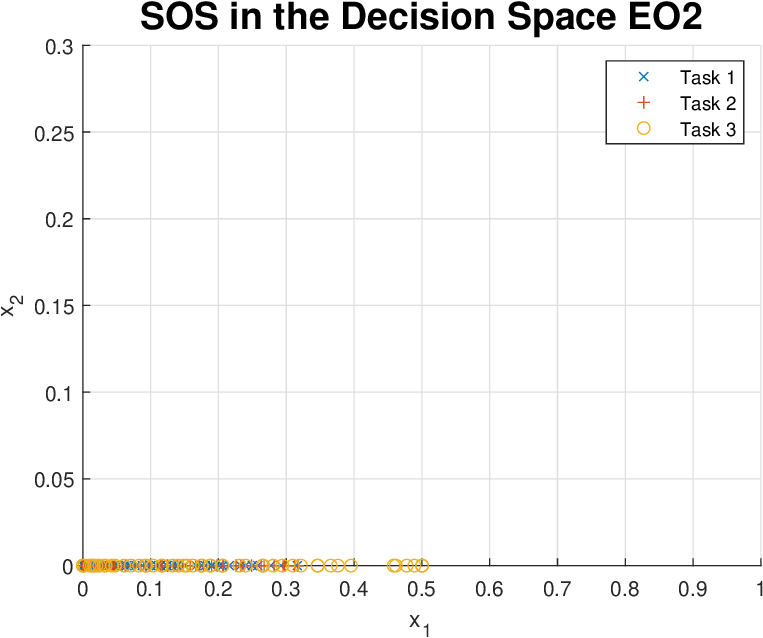}}
        \subfigure[]{\label{re342_dec}\includegraphics[width=0.65\columnwidth]{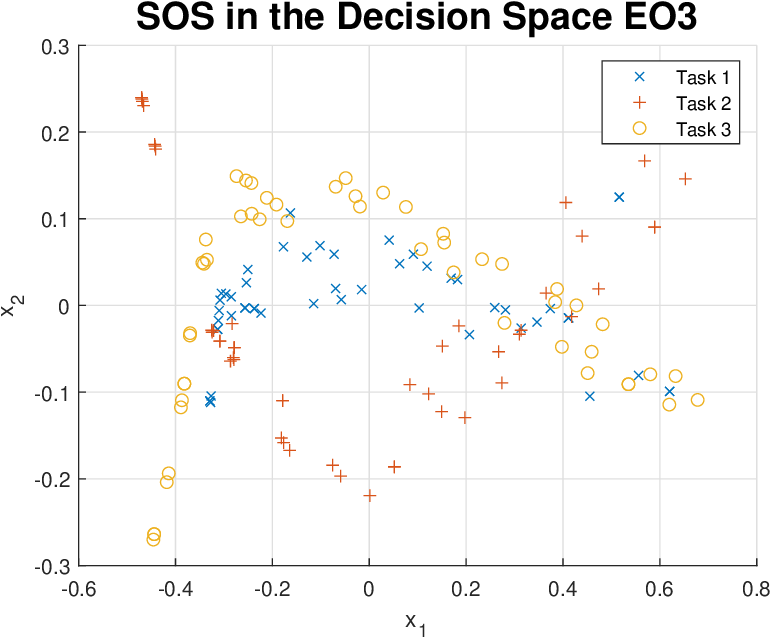}}
        \subfigure[]{\label{im1_dec}\includegraphics[width=0.65\columnwidth]{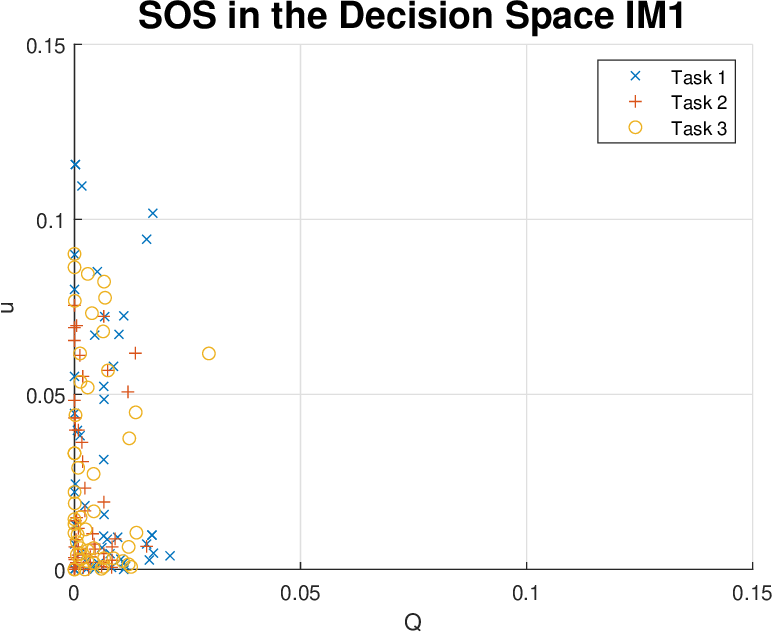}}
		\subfigure[]{\label{im2_dec}\includegraphics[width=0.65\columnwidth]{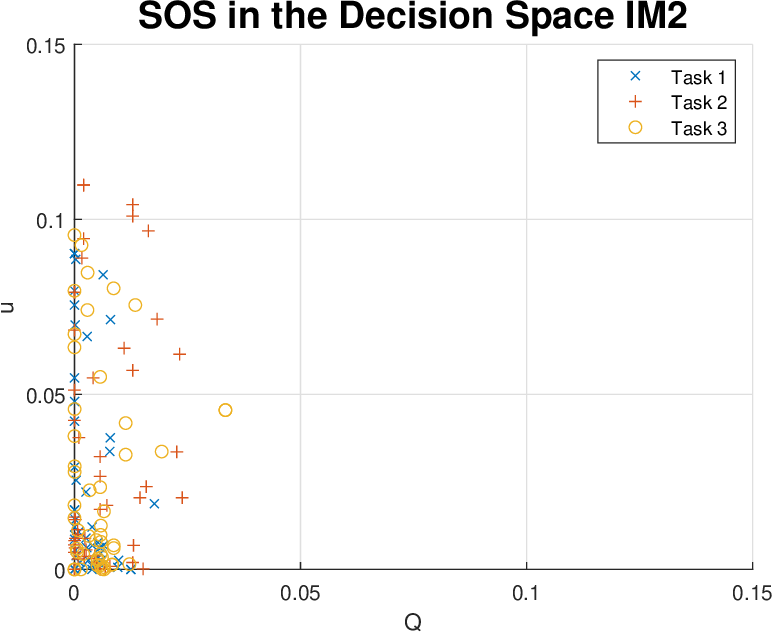}}
        \subfigure[]{\label{im3_dec}\includegraphics[width=0.65\columnwidth]{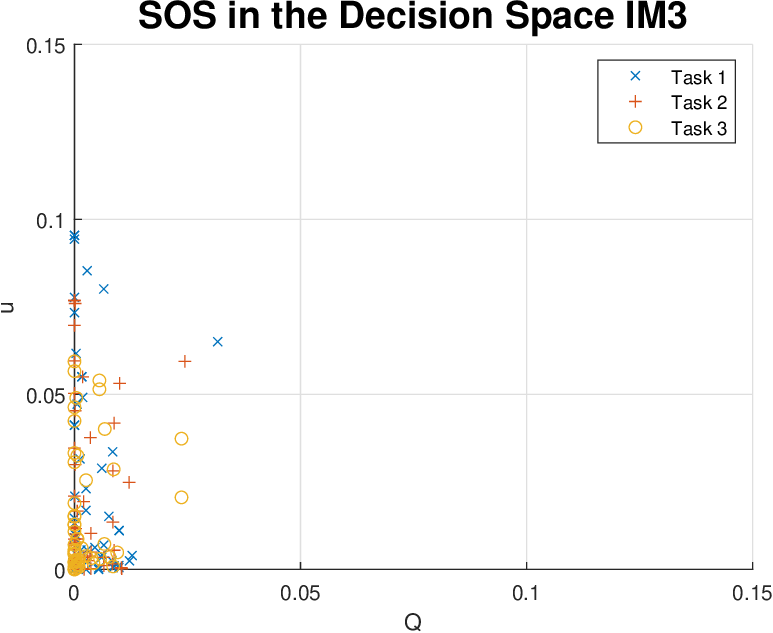}}
        \subfigure[]{\label{mnist_dec}\includegraphics[width=0.65\columnwidth]{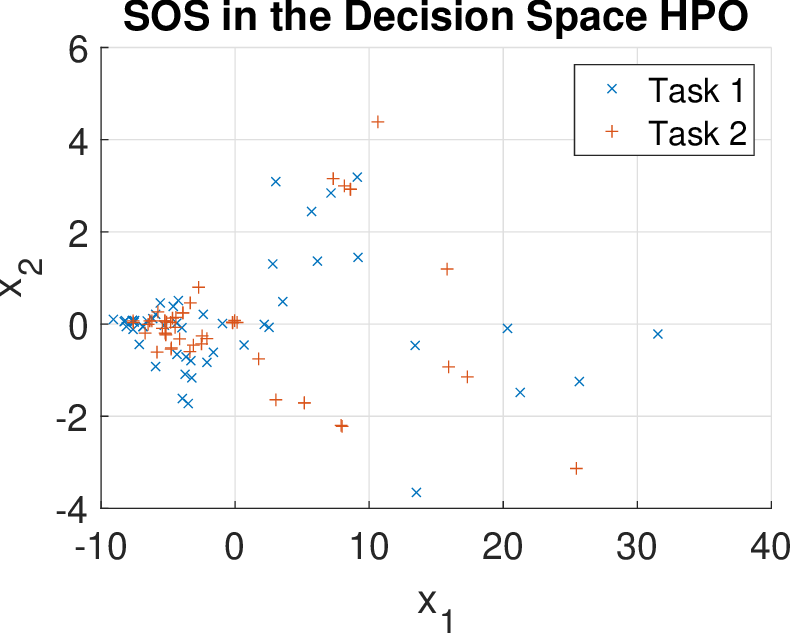}}
		\caption{The SOS shown in the decision space. All of the results are obtained by MO-MFEA. (a) The SOS of EO1. (b) The SOS of EO2. (c) The SOS of EO3. (d) The SOS of IM1. (e) The SOS of IM2. (f) The SOS of IM3. (g) The SOS of HPO.}\label{Fig:SOS_decspace}
	\end{center}
\end{figure*}

\begin{figure*}[!t]
	\begin{center}
		\subfigure[]{\label{re241_obj}\includegraphics[width=0.65\columnwidth]{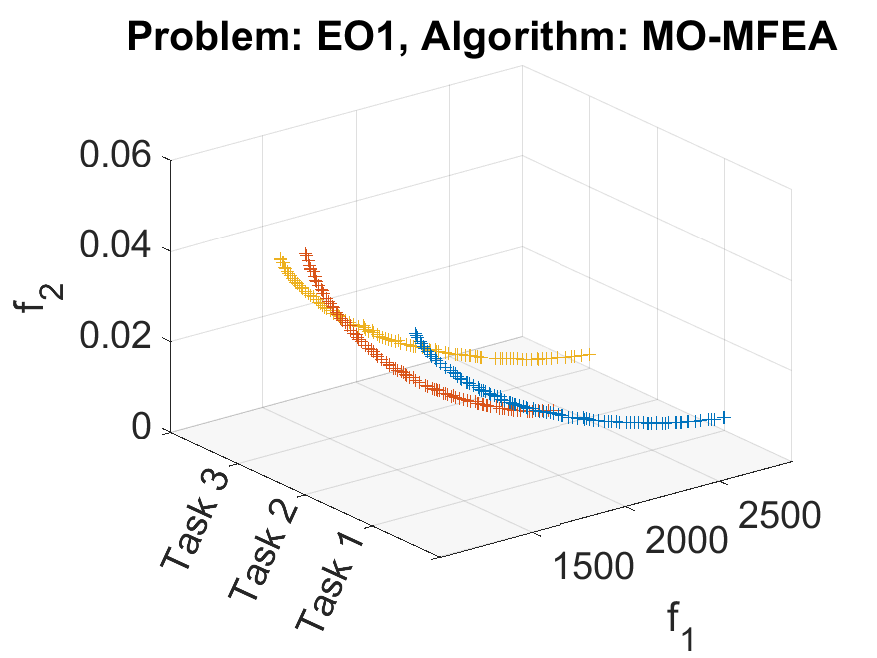}}
		\subfigure[]{\label{re224_obj}\includegraphics[width=0.65\columnwidth]{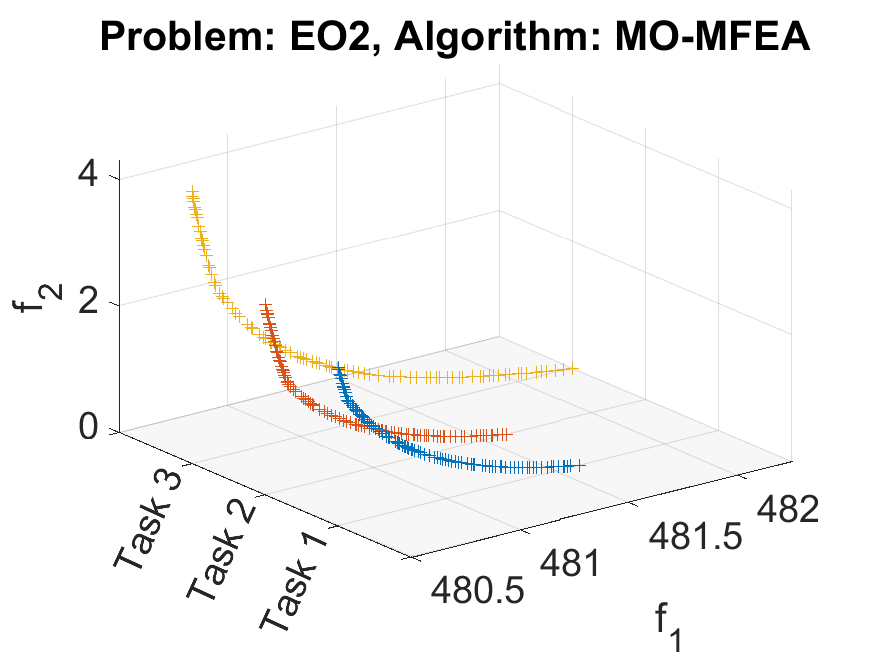}}
        \subfigure[]{\label{re342_obj}\includegraphics[width=0.65\columnwidth]{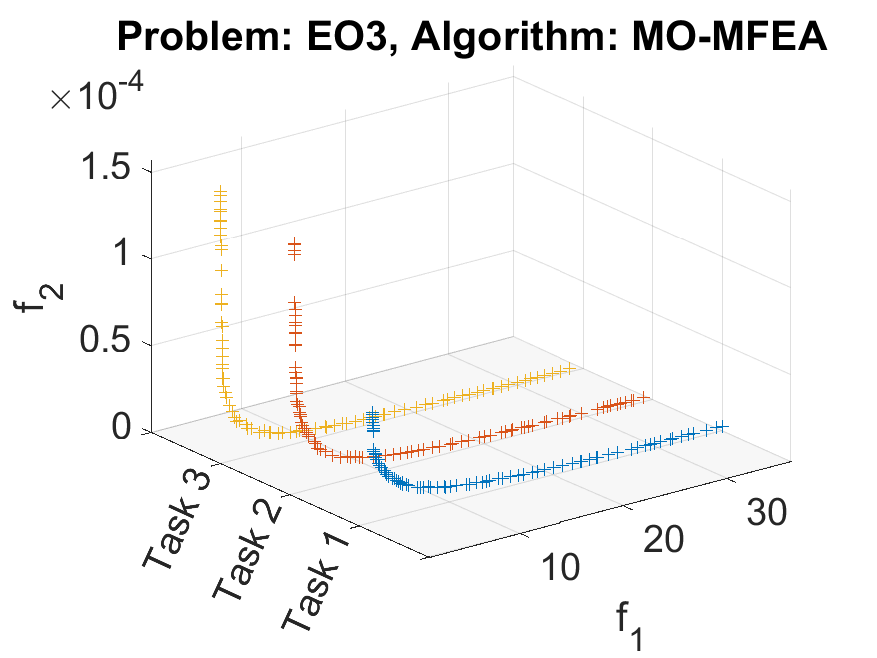}}
        \subfigure[]{\label{im1_obj}\includegraphics[width=0.65\columnwidth]{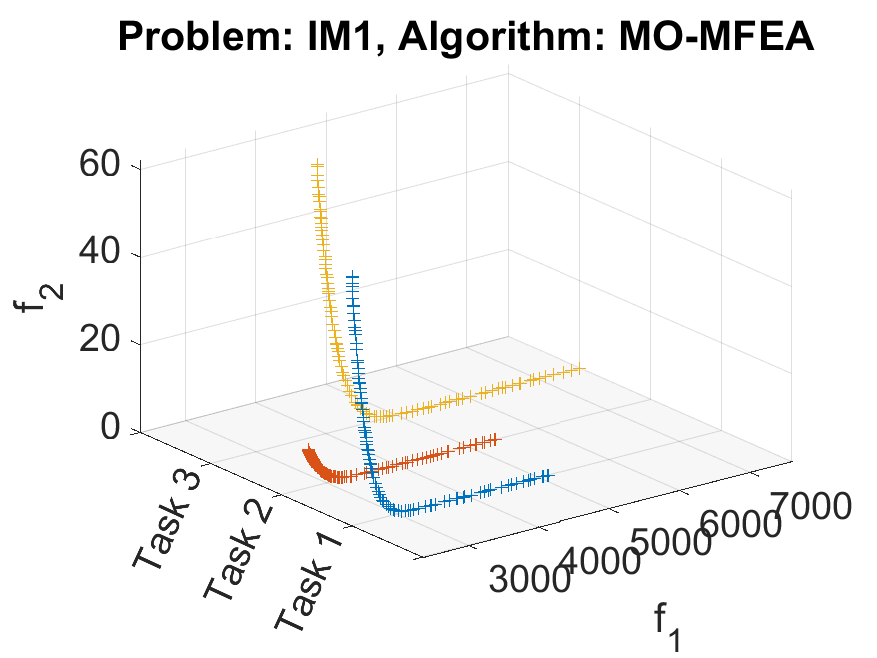}}
		\subfigure[]{\label{im2_obj}\includegraphics[width=0.65\columnwidth]{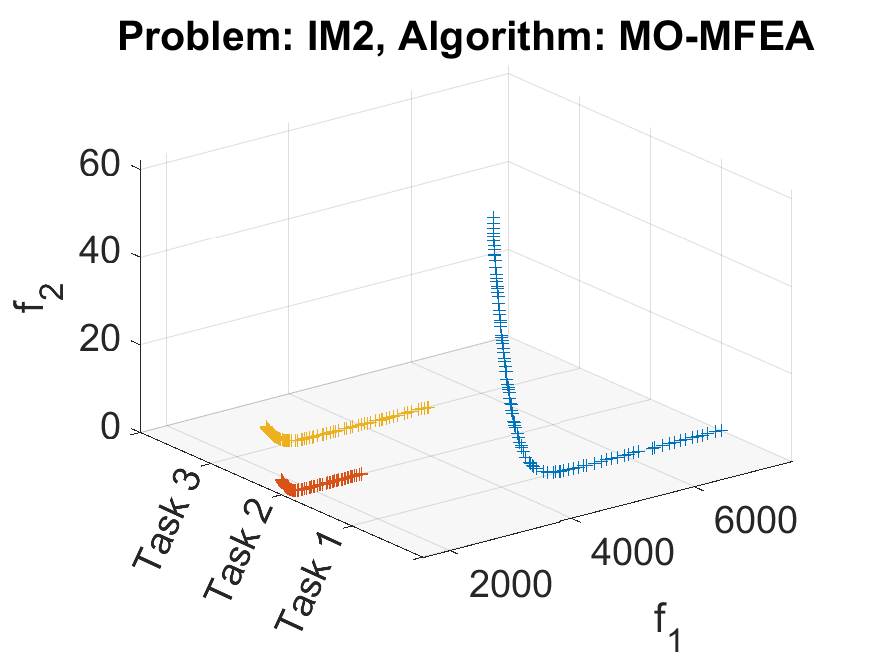}}
        \subfigure[]{\label{im3_obj}\includegraphics[width=0.65\columnwidth]{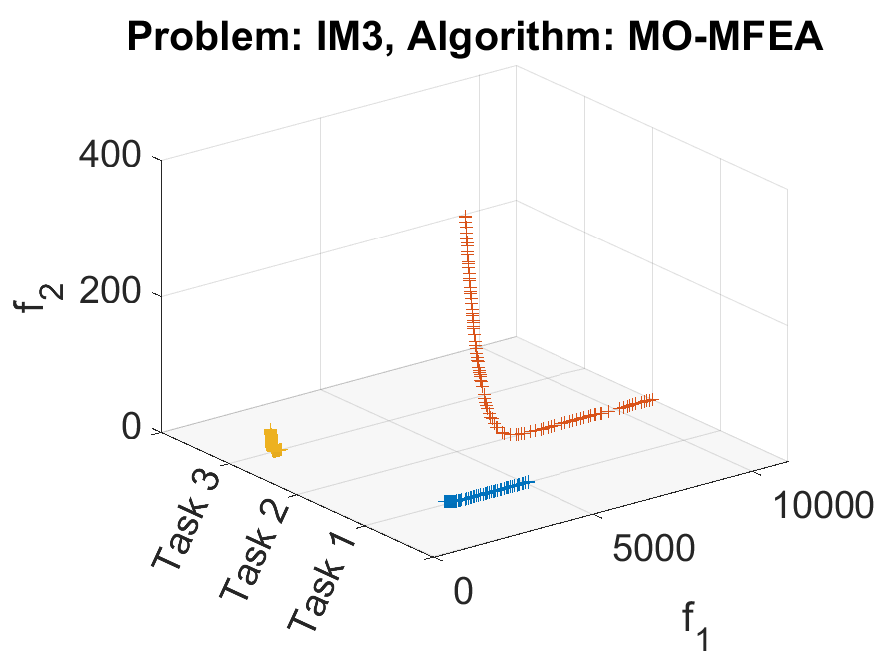}}
        \subfigure[]{\label{mnist_obj}\includegraphics[width=0.65\columnwidth]{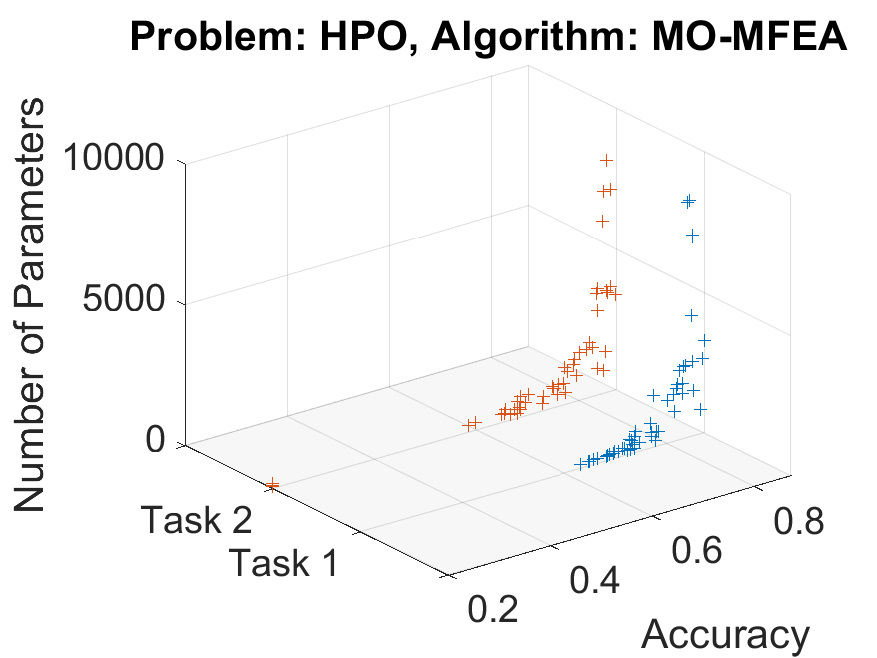}}
		\caption{The SOS shown in the objective space. All of the results are obtained by MO-MFEA. (a) The SOS of EO1. (b) The SOS of EO2. (c) The SOS of EO3. (d) The SOS of IM1. (e) The SOS of IM2. (f) The SOS of IM3. (g) The SOS of HPO.}\label{Fig:SOS_objspace}
	\end{center}
\end{figure*}

\section{Experimental Studies}
\subsection{Experimental Details}
As discussed in Section~\ref{sec2.1}, the generation of the SOS can be modeled as a multitask multiobjective optimization problem, and EMT emerges as a promising technique to accomplish this objective. In this paper, we employ three different EMT algorithms to generate the SOS, thus investigating their capabilities for achieving this purpose. The employed algorithms are listed as follows:
\begin{itemize}
    \item MO-MFEA~\cite{gupta2016multiobjective}: The multiobjective multifactorial evolutionary algorithm.
    
    \item MO-MFEA-II~\cite{bali2020cognizant}: An improved version of the multiobjective multifactorial evolutionary algorithm.
    
    \item EMT-ET~\cite{lin2020effective}: A multiobjective multitasking optimization method with an effective knowledge transfer approach.

\end{itemize}
In additional, we also employ the classical NSGA-II~\cite{deb2002fast} as the baseline. All of the above algorithms are implemented from the MTO-Platform~\cite{Li2023MToP}. The population size of all the algorithms is set to 50. Regarding the engineering design problems and the inventory management problems, the maximum number of evaluations is set to 10000. For the hyperparameter optimization problems, the population size and the maximum number of evaluations are set to 50 and 500, respectively.

\subsection{Results of Different EMT Algorithms}

In this section, we evaluate the quality of the generated SOS by using a metric called \textit{cumulative hypervolume} (CHV). The calculation of the CHV is calculated as follows:
\begin{equation}\label{eqn:sos_result}
\begin{aligned}
CHV & = \sum_{k=1}^{K} \mathcal{HV}( PS'_k ) \\
& = \lambda_{d}( \cup_{y \in PS'_k }[\textbf{y},\textbf{r}_k] )
\end{aligned}
\end{equation}
where $PS'_k$ is the obtained solutions set corresponding to the $k$th task, $\mathcal{HV}(\cdot)$ is the hypervolume~\cite{auger2012hypervolume}, $\lambda_{d}$ being the Lebesgue measure, and $\textbf{r}_k$ is the reference point corresponding to the $k$th task. It's important to note that, before calculating the hypervolume, we normalize the objective function values of all solutions in $PS'_k$ into the region $[0,1]$. Subsequently, each component of the reference point is set to 1. 

Table~\ref{tab:chv} presents the average CHV results obtained by MO-MFEA, MO-MFEA-II, EMT-ET, and NSGA-II over 20 runs. Upon examination, it becomes evident that MO-MFEA-II exhibits superior performance by achieving the best CHV results on six problems. Additionally, another interesting detail is that MO-MFEA-II demonstrates better performance than the classical MO-MFEA. In six out of the seven problems, MO-MFEA-II outperforms MO-MFEA, possibly due to its adaptive transfer parameter estimation strategy. Furthermore, it is noteworthy that EMT methods generally yield better results than single-task algorithms like NSGA-II for most problems, highlighting the effectiveness of the EMT approach.

\subsection{Visualization of the Set of Pareto Sets}
In this subsection, we focus on the visualization of the SOS, aiming to better understand the SOS concept. Firstly, we use MO-MFEA to obtain the SOS of each problem. Then, the SOS is visualized in both the decision and objective spaces.

\subsubsection{Visualization of the Set of Sets in the Decision Space}
We provide a visualization of the SOS in the decision space in Fig.\ref{Fig:SOS_decspace}. In this figure, solutions from different sets are represented by distinct colors. Considering that EO2, IM1, IM2, and IM3 involve two decision variables, we directly showcase the solutions in a \textit{unified } decision space\cite{da2018curbing}, where all decision variables are scaled to the range of $[0,1]$. For the remaining problems, we initially display the solutions in the unified space, subsequently reducing the decision space to two dimensions through principal component analysis~\cite{wold1987principal}. 

The results depicted in Fig.\ref{Fig:SOS_decspace} reveal a notable observation: despite the parameters or environment of the optimization tasks are distinct from each other, the Pareto optimal solutions tend to cluster in similar regions. This characteristic suggests that EMT approaches, with their inherent ability to capture similarities between tasks, may outperform single-task methods on such real-world problems under consideration. This observation aligns with the findings presented in Table \ref{tab:chv} and provides insight into the superior performance of EMT approaches compared to single-task methods.

\begin{table}[]
\centering
\caption{The RMMD Matrices of EO3 and IM1}
\resizebox{8.5cm}{!}{\begin{tabular}{c|ccc||c|ccc}
\hline
\multicolumn{4}{c||}{EO3}   & \multicolumn{4}{c}{IM1}        \\
\hline
      & Task1  & Task2  & Task3  &       & Task1  & Task2  & Task3  \\
\hline
Task1 & 0.0000 & 0.1994 & 0.2604 & Task1 & 0.0000 & 0.0899 & 0.0674  \\
Task2 & 0.1994 & 0.0000 & 0.2017 & Task2 & 0.0899 & 0.0000 & 0.0293  \\
Task3 & 0.2604 & 0.2017 & 0.0000 & Task3 & 0.0674 & 0.0293 & 0.0000  \\
\hline
\end{tabular}}
\end{table}

To further measure the similarity of different Pareto sets, we also develop a measurement called relative mean-minimum distance (RMMD), which is calculated as follows:
\begin{equation}\label{eqn:re241}
\begin{aligned}
\frac{\sum_{i=1}^{N_{k_2}} \min \{ ||\textbf{x}^{(1)*}_{k_1}-\textbf{x}^{(i)*}_{k_2}||_2, \ldots, ||\textbf{x}^{(N_{k_1})*}_{k_1}-\textbf{x}^{(i)*}_{k_2}||_2 \}}{D_{rand}N_{k_2}},
\end{aligned}
\end{equation}
where $\textbf{x}_{k_1}^{(\cdot)*}$ and $\textbf{x}_{k_2}^{(\cdot)*}$ are Pareto solutions corresponding to tasks $k_1$ and $k_2$, respectively, $N_{k_1}$ and $N_{k_2}$ are the number of Pareto solutions corresponding to tasks $k_1$ and $k_2$, respectively, and $D_{rand}$ is obtained by calculating the mean-minimum distance between two randomly sampled population of solutions in the decision space. Generally, a small RMMD indicates a high similarity between Pareto sets, suggesting better performance of EMT. We compute the RMMD matrices for EO3 and IM1, based on the solution sets obtained by MO-MFEA. The comparison reveals that both EO3 and IM1 present values significantly lower than 1, signifying that the Pareto sets corresponding to different tasks exhibit higher similarity and closer distance compared to randomly generated populations. This suggests a strong similarity between the Pareto sets across the tasks. This finding underscores the effectiveness of EMT approaches in such scenarios.

\subsubsection{Visualization of the Set of Sets in the Objective Space}
Fig.~\ref{Fig:SOS_objspace} illustrates the SOS in the objective space. It is crucial to emphasize that the Pareto front of each task constructs the finite SOS. Consequently, in Fig.~\ref{Fig:SOS_objspace}, the SOS manifests as multiple space curves in the objective space. Each of the EO1-EO3 and IM1-IM3 encompasses three sets of solution sets, making the SOS a combination of three space curves. Similarly, the SOS of the HPO is a combination of two space curves, reflecting its two solutions corresponding to two subtasks.




\subsection{An analysis on the Set of Pareto Sets Solutions}
\begin{figure}[!t]
	\begin{center}
		\includegraphics[width=1\columnwidth]{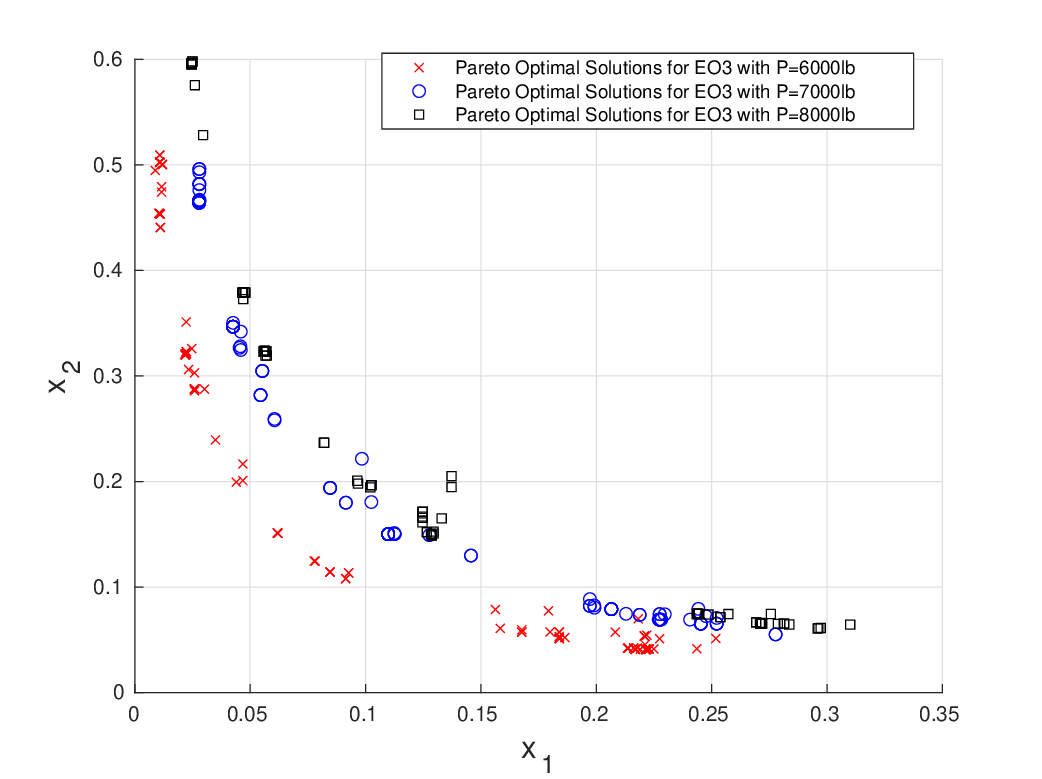}
		\caption{The first two variables of the Pareto sets under $P=6000lb$, $P=7000lb$, and $P=8000lb$.}\label{Fig:discussion}
	\end{center}
\end{figure}

In this subsection, we demonstrate the benefits of arriving at a SOS for different task settings in a single pass using multitask and multiobjective optimization. In particular, by examining the shifts in the trend of Pareto optimal solutions in tandem with variations in the task setting through the SOS attained provide an opportunity for deep understanding of the optimal designs and inherent trade-offs in objectives, thereby empowering users or engineers to grasp the nuanced effects of their design choices. Using EO3 as an example, we analyze the three Pareto sets corresponding to $P = 6000lb$, $P = 7000lb$, and $P = 8000lb$ ($L$ and $E$ are set to $14in$ and $3.00E+07psi$ respectively) in one pass obtained using MO-MFEA-II for multitask multiobjective optimization. Through visual analysis of the first two decision variables of the solutions across different Pareto sets in Fig.~\ref{Fig:discussion}, we observe that as $P$ increases, the values of these decision variables, namely the length of the weld seam and the welding melt depth, tend to slightly increase. This insight enables engineers to infer that when targeting Pareto optimal solutions for higher load forces, it would be prudent to consider slightly increasing the values of these decision variables based on the solutions obtained for lower load forces. This observation aligns with the physical characteristics of EO3, wherein improving the length of the weld seam and welding melt depth enhances the stability of the beam under larger load forces. This systematic analysis facilitates engineers in developing a deeper understanding of the complexities inherent in such engineering design problems.

\section{Conclusion}
As a novel concept, the SOS not only demonstrates its potential in the field of machine learning but also proves valuable in various domains such as engineering and management science. Modeling the generation of the SOS as a multitask multiobjective optimization problem becomes natural when considering finite sets of solution sets. In this context, EMT emerges as an effective methodology for handling the SOS. Unlike previous research primarily focused on investigating the performance of EMT methods using benchmark problems, this paper delves into exploring these methods' capabilities in generating SOS for real-world problems. We have studied three categories of real-world problems, encompassing sets of engineering design problems, inventory management problems, and hyperparameter optimization problems. Five EMT algorithms are utilized to generate the SOS. The experimental results visualize the SOS in both the objective spaces and the decision space and demonstrate the effectiveness of current EMT algorithms in generating SOS for real-world problems. Last but not least, we show that analyzing the changes in the trend of Pareto optimal designs in correlation with variations in the task setting through the SOS solutions offers valuable insights into the dynamic interplay between design solutions and their performance in different contexts. This analytical approach serves to enhance users' understanding of how Pareto optimal designs respond to varying task settings, shedding light on their adaptability and effectiveness under diverse settings such as environmental conditions.

\section*{Acknowledgment}
This research is partly supported by the National Research Foundation, Singapore and DSO National Laboratories under the AI Singapore Programme (AISG Award No.: AISG2-GC-2023-010, ``Design Beyond What You Know": Material-Informed Differential Generative AI (MIDGAI) for Light-Weight High-Entropy Alloys and Multi-functional Composites (Stage 1a)", the  the Honda Research Institute Europe GmbH, and the College of Computing and Data Science, Nanyang Technological University.

\bibliographystyle{IEEEtran}
\bibliography{ref}


\end{document}